\documentclass{article}

\PassOptionsToPackage{numbers, compress}{natbib}

\usepackage[preprint]{nips_2018}

\usepackage{color}
\usepackage{cuted}
\usepackage{fullpage,enumitem,amsmath,amssymb,graphicx,float}

\usepackage[utf8]{inputenc} 
\usepackage[T1]{fontenc}    
\usepackage{graphicx}
\usepackage{hyperref}       
\usepackage{url}            
\usepackage{booktabs}       
\usepackage{amsfonts}       
\usepackage{nicefrac}       
\usepackage{microtype}      
\usepackage{booktabs}
\usepackage{multirow}
\usepackage{amsmath}
\usepackage{appendix}
\usepackage{caption}
\usepackage[table,xcdraw]{xcolor}
\usepackage{subcaption}
\newcommand{\pvocab}{\text{P}_{\text{vocab}}}
\newcommand{\pfinal}{\text{P}_{\text{final}}}
\title{\large{Neural Abstractive Text Summarization and Fake News Detection}}

\author{
  Soheil Esmaeilzadeh \\
   \texttt{soes@stanford.edu} \\
    Stanford University, CA  \\
   \And
   Gao Xian Peh \\
    \texttt{gxpeh@stanford.edu} \\
    Stanford University, CA  \\
   \And
  Angela Xu \\
     \texttt{aaxu@stanford.edu} \\
   Stanford University, CA  \\
}

\begin{document}
\maketitle
\begin{abstract}
In this work, we study abstractive text summarization by exploring different models such as LSTM-encoder-decoder with attention, pointer-generator networks, coverage mechanisms, and transformers. Upon extensive and careful hyperparameter tuning we compare the proposed architectures against each other for the abstractive text summarization task. Finally, as an extension of our work, we apply our text summarization model as a feature extractor for a fake news detection task where the news articles prior to classification will be summarized and the results are compared against the classification using only the original news text. 
\vspace{3.5pt}
\\
{\textbf{keywords:}} LSTM, encoder-deconder, abstractive text summarization, pointer-generator, coverage mechanism, transformers, fake news detection
\end{abstract}

\section{Introduction}
Pattern recognition and data understanding has been the topic of research in multiple deep learning tasks such computer vision and natural language processing \cite{esmaeilzadeh_alz3,esmaeilzadeh_alz,esmaeilzadeh_alz2,Cheng2019,Liu2018}. In the natural language processing area, understanding the content and main idea of a text and summarizing a corpus is of great importance. In simple words, text summarization is the task of creating a summary for a large piece of text. Generating \textit{meaningful} summaries of long texts is of great importance in many different areas such as medical, educational, media, social, and etc., where the summary needs to contain the main contextual aspects of the text while reducing the amount of unnecessary information. 

In general, text summarization can be classified into two main groups: \textit{extractive summarization} and \textit{abstractive summarization} \citep{Allahyari}. Extractive summarization creates summaries by synthesizing salient phrases from the full text verbatim \citep{Dorr2007,Nallapati2016}, however, abstractive summarization creates an internal semantic representation of the text. Unlike extractive summarization which concatenates sentences taken explicitly from the source text, abstractive text summarization paraphrases the text in a way that it is closer to the human's style of summarization and this makes abstractive text summarization a challenging yet preferable approach \citep{Khatri2018,Gao}. 

Decent quality summaries using abstractive approaches were only obtained in the past few years by applying the sequence-to-sequence endoder-decoder architectures with attention mechanisms common in machine translation tasks to summarization \citep{Bahdanau,Nallapati2016a} however only focused on short input texts. Subsequent works attempted to perform the abstractive summarization task on longer input texts, however, appearance of unknown words and repetitions adversely affected the outcome of the summarization tasks \citep{Moritz}.

In this work, we focus on abstractive text summarization as a more robust approach compared to its counterpart (i.e. extractive summarization) and explore recent advancements in the state-of-the-art natural language models for abstractive text summarization. The input of our natural language model is a single document or article and the output of it is a combination of a few sentences that summarize the content of the input document in a meaningful manner. In addition to the main goal of this work, after exploring the natural language models for abstractive text summarization, we use the summarization model as a feature building module for fake news detection and news headline generation, and show the effect of summarization on fake news detection.

\section{Approaches} \label{sec_approaches}
\begin{figure}[!ht]
\centering
\includegraphics[width=0.9\textwidth]{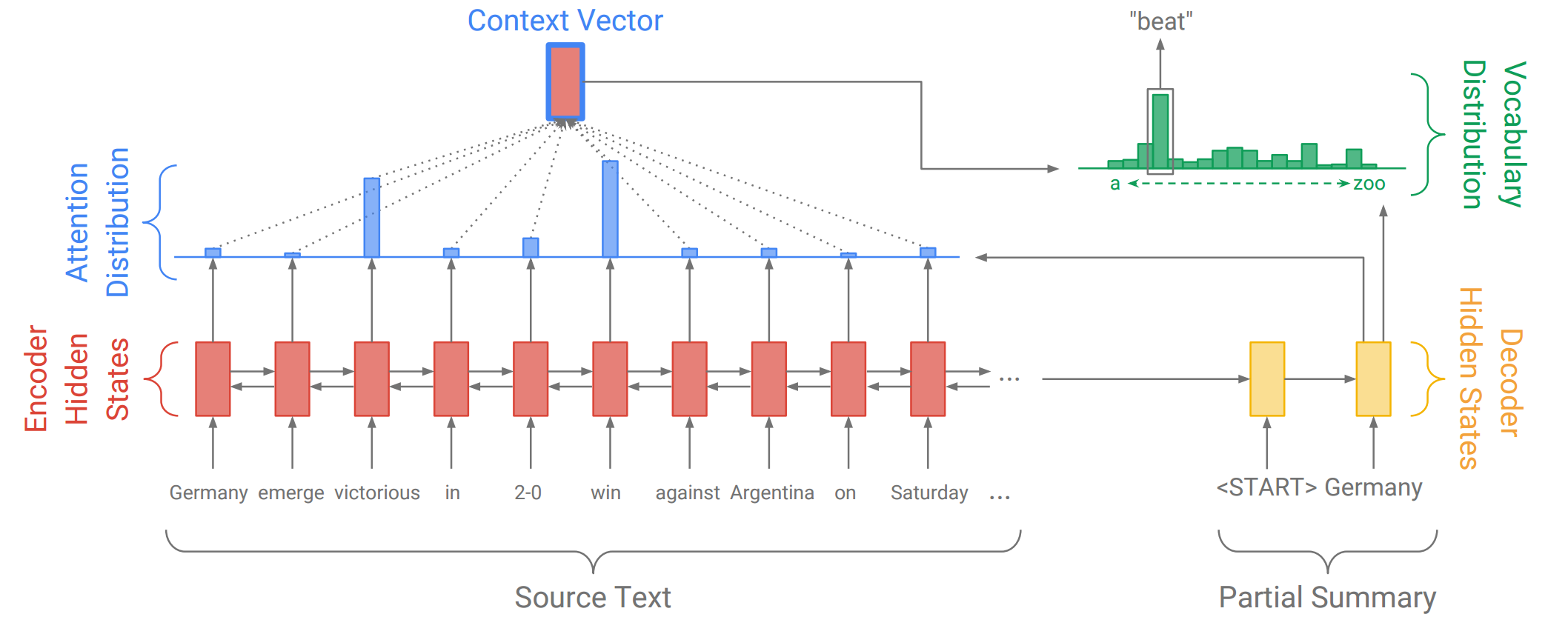}
	\caption{Baseline sequence-to-sequence model's architecture with attention \citep{See2013}}
	\label{fig_baseline}
\end{figure}
\subsection{Baseline Model} \label{sec_baseline}
%

%
In this work, as the baseline model we consider an LSTM Encoder-Decoder architecture with attention as shown in Figure \ref{fig_baseline}.  

\textbf{Sequence-to-Sequence Encoder-Decoder:} The sequence-to-sequence framework consists of a Recurrent Neural Network (RNN) encoder and an RNN decoder. The RNN encoder as a single-layer bidirectional Long-short-term-memory (LSTM) unit reads in the input sequence token by token and produces a sequence of encoder's hidden states $h_i$ that encode or represent the input. The RNN decoder as a single-layer unidirectional LSTM generates the decoder's hidden states $s_t$ one by one which produces the output sequence as the summary.  

\textbf{Attention Mechanism:} In the attention mechanism, an attention distribution $a^t$ is calculated as a probability distribution over the words in the source text that helps the decoder decide which source words to concentrate on when it generates the next word. The attention distribution $a^t$ is calculated for each decoder timestep $t$ as:
\begin{subequations}
\begin{equation}
e^t_i = v^T \text{tanh}(W_hh_i + Ws_t + b_{attn}),
\label{Eq_et}
\end{equation} 
\vspace{-10pt}
\begin{equation}
a^t = \text{softmax}(e^t),
\end{equation}
\end{subequations}
where $v$, $W_h$, $W_s$, $b_{attn}$ are learnable parameters. On each decoder's step, attention weights $a^t_i$, which are part of the $a^t$ distribution for the source words are computed.  An attention weight represents the amount of attention that should be paid to a certain source word in order to generate an output word (decoder state) in the decoder. The attention distribution is used to compute a weighted sum of the encoder hidden states, known as the context vector $h^*_t$, which represents what has been read from the source for this step, and can be calculated as:
\begin{equation}
h^*_t = \sum_i a^t_i h_i\,.
\end{equation}
The context vector along with the decoder's state are then used to calculate the vocabulary distribution $\pvocab$, which provides a final distribution for predicting words $w$ as:
\begin{subequations}
\begin{equation}
\pvocab = \text{softmax}(V^\prime(V[s_t, h^*_t] + b) + b^\prime),
\end{equation}
\vspace{-10pt}
\begin{equation}
\text{P}(w) = \pvocab(w),
\end{equation}
\end{subequations}
where $V$, $V^\prime$, $b$, and $b^\prime$ are learnable parameters. Subsequently, we calculate the loss for the timestep $t$ as the negative log-likelihood of the target word $w^*_t$ as: 
\begin{equation} \label{eq_losst}
    \text{loss}_t = -\log \text{P}(w^*_t)\,.
\end{equation}
The overall loss for the whole sequence is the average of the loss at each time step (i.e. $\text{loss}_t$) as: \\
\begin{equation}
  \text{loss} = \frac{1}{T} \sum_{t=0}^T \text{loss}_t\,.  
\end{equation}
\textbf{Baseline Model's Problems:} Some problems are associated with the baseline model proposed in section \ref{sec_baseline}. One problem is the model's tendency to reproduce factual details inaccurately, this happens specially when an uncommon word that exists in the vocabulary is replaced with a more common word. Another problem with the baseline model is that during summary generation it repeats the already generated parts of the summary. Lastly, the baseline is unable to handle out-of-vocabulary words (OOV). In general, it is hard for the sequence-to-sequence-with-attention model to copy source words as well as to retain longer-term information in the decoder state, which leads to the aforementioned issues. \citet{See2013} proposed a so called \textit{pointer-generator} network that also includes a coverage mechanism in order to address these problems by combining both context extraction (pointing) and context abstraction (generating). We revisit the model proposed by \citet{See2013} in the following and as well compare it with a transformer based model proposed by \cite{Vaswani2017} for machine translation tasks, and finally use it as a feature generation mechanism for fake news classification.
\subsection{Pointer-Generator Network}
\begin{figure}[!ht]
\centering 
\includegraphics[width=0.9\textwidth]{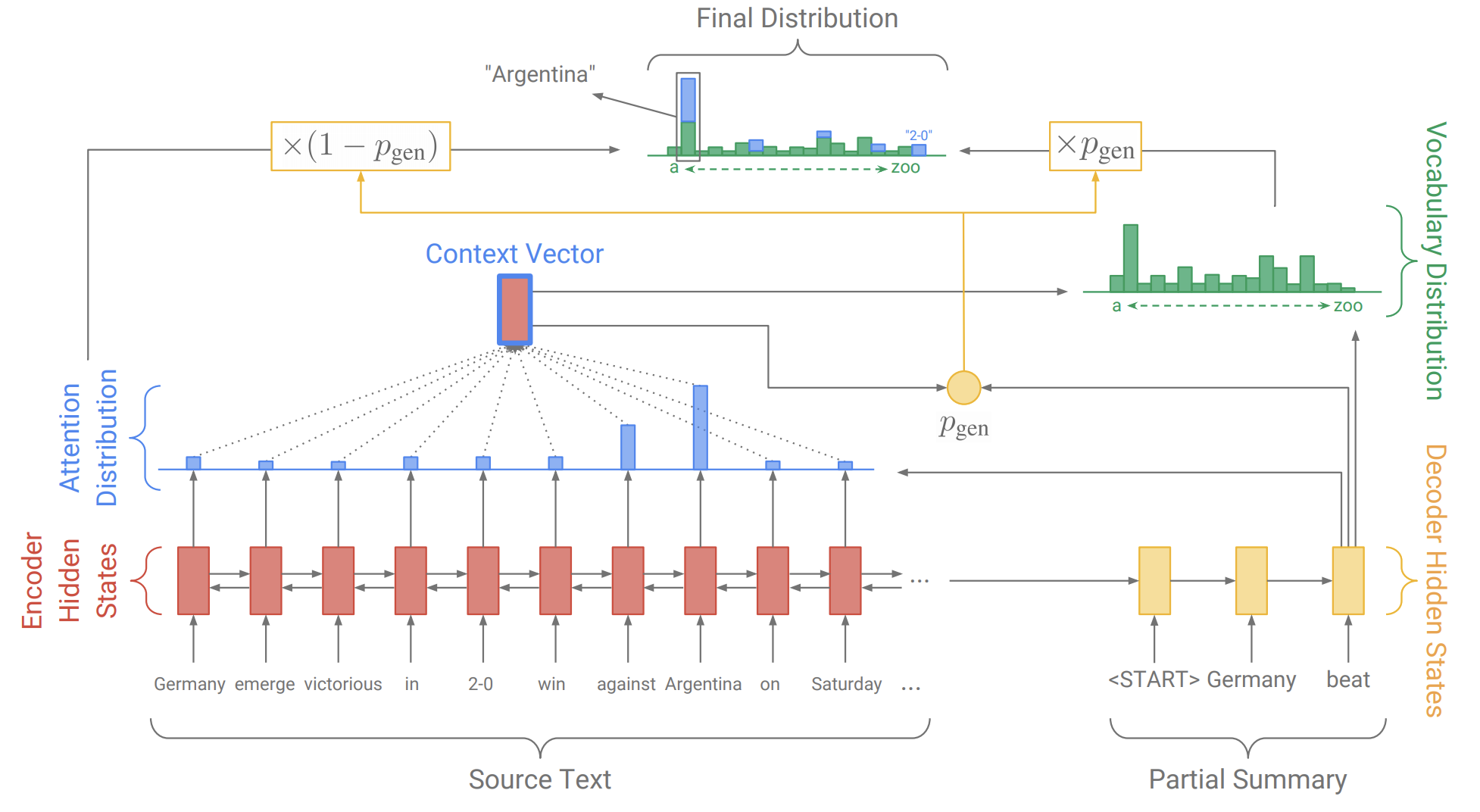}
	\caption{Pointer-generator model's architecture \citep{See2013}}
	\label{fig_pointerGenerator}
\end{figure}
\textbf{Pointer-Generator Mechanism:} Pointer-generator is a hybrid network that chooses during training and test whether to copy words from the source via pointing or to generate words from a fixed vocabulary set. 
Figure \ref{fig_pointerGenerator} shows the architecture for the pointer-generator mechanism where the decoder part is modified compared to Figure \ref{fig_baseline}. In Figure \ref{fig_baseline}, the baseline model, only an attention distribution and a vocabulary distribution are calculated. However, here in the pointer-generator network a generation probability $p_{\text{gen}}$, which is a scalar value between 0 and 1 is also calculated which represents the probability of generating a word from the vocabulary, versus copying a word from the source text. The generation probability $p_{\text{gen}}$ weights and combines the vocabulary distribution $\pvocab$ (used for generating) and the attention distribution $a$ (used for pointing to source words $\omega_i$) into the final distribution $\pfinal$ as:
\begin{equation}
\label{eq_pfinal}
 \pfinal(w) = p_{\text{gen}}\pvocab(w) + (1-p_{\text{gen}})\sum_{i:w_i = w} a_i\,.
\end{equation}
Based on Equation \eqref{eq_pfinal}, the probability of producing word $\omega$ is equal to the probability of generating it from the vocabulary multiplied by the generation probability plus the probability of pointing to it anywhere it appears in the source text multiplied by the copying probability. Compared to the LSTM Encoder-Decoder model with attention as baseline in section \ref{sec_baseline}, the pointer-generator network makes it easy to copy words from the source text by putting sufficiently large attention on the relevant word. It also is able to copy out-of-vocabulary words from the source text, enabling the model to handle unseen words while allowing to use a smaller vocabulary, leading to less computation and storage space. The pointer-generator model is also faster to train, as it requires fewer training iterations to achieve the same performance as the baseline model in section \ref{sec_baseline}. 

\textbf{Coverage Mechanism:} To reduce the repetition during summarization as a common issue with sequence-to-sequence models mentioned in section \ref{sec_baseline}, we apply the coverage mechanism, first proposed by \cite{Tu2016} and adapted by \cite{See2013}.  The coverage mechanism keeps track of a coverage vector, computed as the sum of attention distributions over previous decoder time steps.  This coverage vector is incorporated into the attention mechanism and represents the degree of coverage that words in the source text have received from the attention mechanism so far. Thus, by maintaining this coverage vector, which represents a cumulative attention, the model avoids attending to any word that has already been covered and used for summarization.\\
On each timestep $t$ of the decoder, the coverage vector $c^t$ is the sum of all the attention distributions ${a^t}^\prime$ so far as: 
\begin{equation}
   c^t = \sum_{t^{\prime}=0}^{t-1} a^{t^{\prime}}\,.
\end{equation}
This coverage vector also contributes to computing the attention mechanism described in the previous section, so that Equation \eqref{Eq_et} becomes:
\begin{equation}
    e^t_i = v^T \text{tanh}(W_hh_i + Ws_t + w_c c^t_i + b_{attn})\,.
\end{equation}
Intuitively, this informs the attention mechanism's current timestep about the previous attention information which is captured in $c^t$, thus preventing repeated attention to the same source words.   
To further discourage repetition, \citet{See2013} penalizes repeatedly attending to the same parts of the source text by defining a coverage loss and adding it to the primary loss function in Equation \eqref{eq_losst}. This extra coverage loss term penalizes any overlap between the coverage vector $c^t$ and the new attention distribution $a^t$ as:
\begin{equation}
    \text{covloss}_t = \sum_i \text{min}(a_i^t, c_i^t)\,.
\end{equation}
Finally the total loss becomes: $\text{loss} = \frac{1}{T} \sum_{t=0}^T (\text{loss}_t + \text{covloss}_t)$. For the aforementioned models we have consulted the Gihub repositories referenced at the end of this report.

\begin{figure}
     \centering
     \begin{subfigure}[b]{0.49\textwidth}
         \centering
         \includegraphics[scale = 1.75]{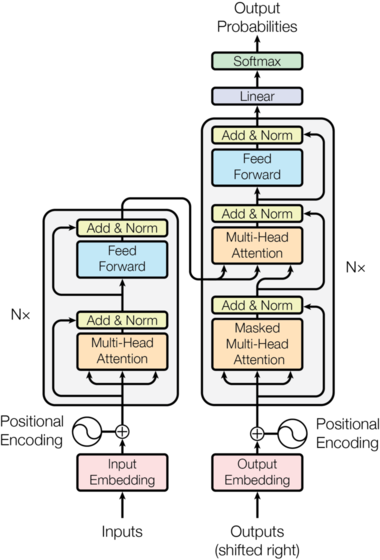}
         \caption{}
         \label{fig_transformer1}
     \end{subfigure}
     \hfill
     \begin{subfigure}[b]{0.49\textwidth}
         \centering
         \includegraphics[scale = 0.47]{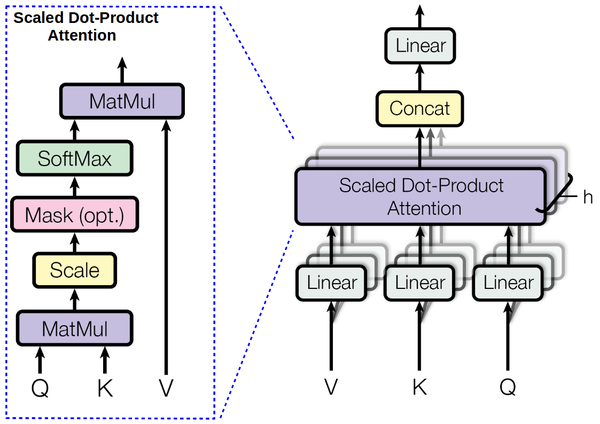}
         \caption{}
         \label{fig_transformer2}
     \end{subfigure}
     \caption{(a). The Transformer - model architecture, (b). (left) Scaled Dot-Product Attention. (right) Multi-Head Attention consists of several attention layers running in parallel - \citep{Vaswani2017}}
     \label{fig_transformer}
\end{figure}
\subsection{Transformers}
In this part, we revisit the \textit{transformers network} proposed by \citet{Vaswani2017} for machine translation, and investigate its performance on abstractive text summarization on our dataset. In the transformer model, the encoder maps an input sequence of symbol representations as $\mathbf{x} = (x_1, ..., x_n)$ to a sequence of continuous representations as $\mathbf{z} = (z_1, ..., z_n)$. Given $\textbf{z}$, the decoder then generates an output sequence as $\mathbf{y} = (y_1, ..., y_n)$ of symbols one element at a time. At each step the model is auto-regressive, consuming the previously generated symbols as additional input when generating the next. The Transformer follows this overall architecture using stacked self-attention and point-wise fully connected layers for both the encoder and decoder, shown in the left and right halves of Figure \ref{fig_transformer}, respectively. 
The encoder part of this architecture is mainly a stack of some identical layers where each one has two sublayers. The first is a multi-head self-attention mechanism, and the second is a simple, position wise fully connected feed-forward network. 
The decoder is also composed of a stack of identical layers. In addition to the two sub-layers in each encoder layer, the decoder inserts a third sub-layer, which performs multi-head attention over the output of the encoder stack. 
In the transformer architecture a variation of attention mechanism called \textit{Scaled Dot-Product Attention} is used where the input consists of queries and keys of dimension $d_k$, and values of dimension $d_v$. The dot products of the query with all keys is calculated, then divided by $\sqrt{d_k}$, the result goes through a softmax function to obtain the weights on the values. In practice the attention function is computed on a set of queries simultaneously, packed together into a matrix $Q$. The keys and values are also packed together into matrices $K$ and $V$, where the matrix of output can be calculated as:
\begin{equation}
    \text{Attention}(Q,K,V) = \text{softmax}(\frac{QK^T}{\sqrt{d_k}})V\,.
\end{equation}
In the proposed transformer model by \cite{Vaswani2017} instead of performing a single attention function they linearly project the queries, keys, and values different times with different learned linear projections and that way they build a \textit{multi-head attention}. On each of the projected versions of queries, keys, and values they then perform the attention function in parallel, yielding multi-dimensional output values which are concatenated and once again projected (Figure \ref{fig_transformer2}). For the transformer model we have consulted the Gihub repositories referenced at the end of this report.
\section{Experiments}
\subsection{Dataset Overview \& Preprocessing}
To train our summarization models, we use the CNN-Dailymail dataset, a collection of news articles and interviews that have been published on the two popular news websites CNN.com and Dailymail.com. Like the common styles on newspapers and journals, each article contains 3-4 highlighted sections that together form the summary of the whole article. The raw dataset includes the text contents of web-pages saved in separate HTML files \citep{Chen,Wang}. We use the CNN and Dailymail dataset provided by DeepMind. Our dataset is split in 92\%, 4.2\%, 3.8\% between training, dev, and test set respectively leading to 287,200 training pairs, 13,360 validation pairs, and 11,400 test pairs. There is an average of 781 tokens per news article. Each reference summary contains 3.75 sentences and 56 tokens on average. \\
We preprocess the dataset and convert the characters all to lower case. We use the Stanford CoreNLP library to tokenize the input articles and their corresponding reference summaries and to add paragraph and sentence start and end markers as $<p>,\,</p>$ and $<s>,\,</s>$ respectively. In addition, we have tried limiting our vocabulary size to 150k and 50k. 
\subsection{Evaluation Metric}
We evaluate our models with the standard ROUGE (Recall-Oriented Understudy for Gisting Evaluation) score, a measure of the amount of overlap between the system-generated and the reference summaries (\cite{Lin2001}). We report the F1, precision, and recall scores for ROUGE-1, ROUGE-2, and ROUGE-L, which measure respectively the word-overlap, bigram-overlap, and longest common sequence between the system-generated and reference summaries. The ROUGE recall and precision for summarization task can be calculated as:
\begin{subequations}
\begin{equation} \label{eq_rouge1}
\text{ROUGE recall} = \frac{\text{number of overlapping words}}{\text{total words in reference summary}}\,,
\end{equation}    
\begin{equation}\label{eq_rouge2}
\text{ROUGE precision} = \frac{\text{number of overlapping words}}{\text{total words in system summary}}\,,
\end{equation}
\end{subequations}
where the system summary refers to the summary generated by a summarization model. Using precision, it's possible to measure
essentially how much of the system summary was in fact relevant or needed, and using recall ROUGE it's possible to measure how much of the reference summary is the summarization model generating. In terms of measuring the overlapping words in Equations \eqref{eq_rouge1} and \eqref{eq_rouge2}, considering the overlap of \textit{unigrams}, or \textit{bigrams}, or \textit{longest common sequence} leads to ROUGE-1, ROUGE-2, and ROUGE-L scores respectively for precision and recall.
\subsection{Experimental Details \& Results and Analysis} \label{sec_results}
\textbf{Text Summarization}

In this work, we investigate the performance of the summarization models presented in section \ref{sec_approaches}  namely: (1). LSTM encoder-decoder with only attention mechanism (baseline), (2). LSTM encoder-decoder with attention and pointer-generator mechanisms, (3). LSTM encoder-decoder with attention, pointer-generator, and coverage mechanisms, and (4). transformers. Table \ref{tab_PGModels} shows the ROUGE-1, ROUGE-2, and ROUGE-L scores for the four different models that have been trained on the summarization dataset. We have trained the models upon hyperparameter tuning using Adagrad optimizer for 340 iterations (19 epochs). Our training results outperform the similar ones presented by \cite{See2013} for cases [1] and [2], and are very close in case [3].
\begin{table}[!ht]
\centering
\renewcommand{\arraystretch}{1.1}
\resizebox{0.9\columnwidth}{!}{%
\begin{tabular}{|c|c|c|c|c|c|c|c|c|c|}
\hline
\multirow{3}{*}{{\textbf{Model}}}      & \multicolumn{9}{c|}{\textbf{{ROUGE}}}                   \\ \cline{2-10} 
                                     & \multicolumn{3}{c|}{\textbf{1}}  & \multicolumn{3}{c|}{\textbf{2}}   & \multicolumn{3}{c|}{\textbf{L}}                    \\ \cline{2-10} 
                                     & \textbf{F1} & \textbf{Precision} & \textbf{Recall} & \textbf{F1} & \textbf{Precision} & \textbf{Recall} & \textbf{F1} & \textbf{Precision} & \textbf{Recall} \\ \hline
                                     
{[1]} & 

35.68	& 44.07	&31.95	&14.21	&17.66	&12.87	&30.56	&41.02	&29.67 \\ \hline

{[2]} & 
38.47	&43.02	&36.98	&16.33&	18.68	&15.94	&33.37&	39.60&	33.99\\ \hline

{[3]} & 
38.97 & 42.71 & 38.21 &16.81 & 18.12 & 16.22& 35.41 &38.63 & 35.04\\ \hline

{[4]} & 
36.55 & 43.33 & 34.50 &15.21 & 17.92 & 13.89& 31.19 &40.38 & 31.54\\ \hline
\end{tabular}
}

\caption{[1]. LSTM encoder-decoder with only attention mechanism (baseline), [2]. LSTM encoder-decoder with attention and pointer-generator mechanisms, [3]. LSTM encoder-decoder with attention, pointer-generator, and coverage mechanisms, and [4]. transformers}
\label{tab_PGModels}
\end{table}
\vspace{-25 pt}
\begin{figure}[!ht]
\centering 
\includegraphics[width=0.55\textwidth]{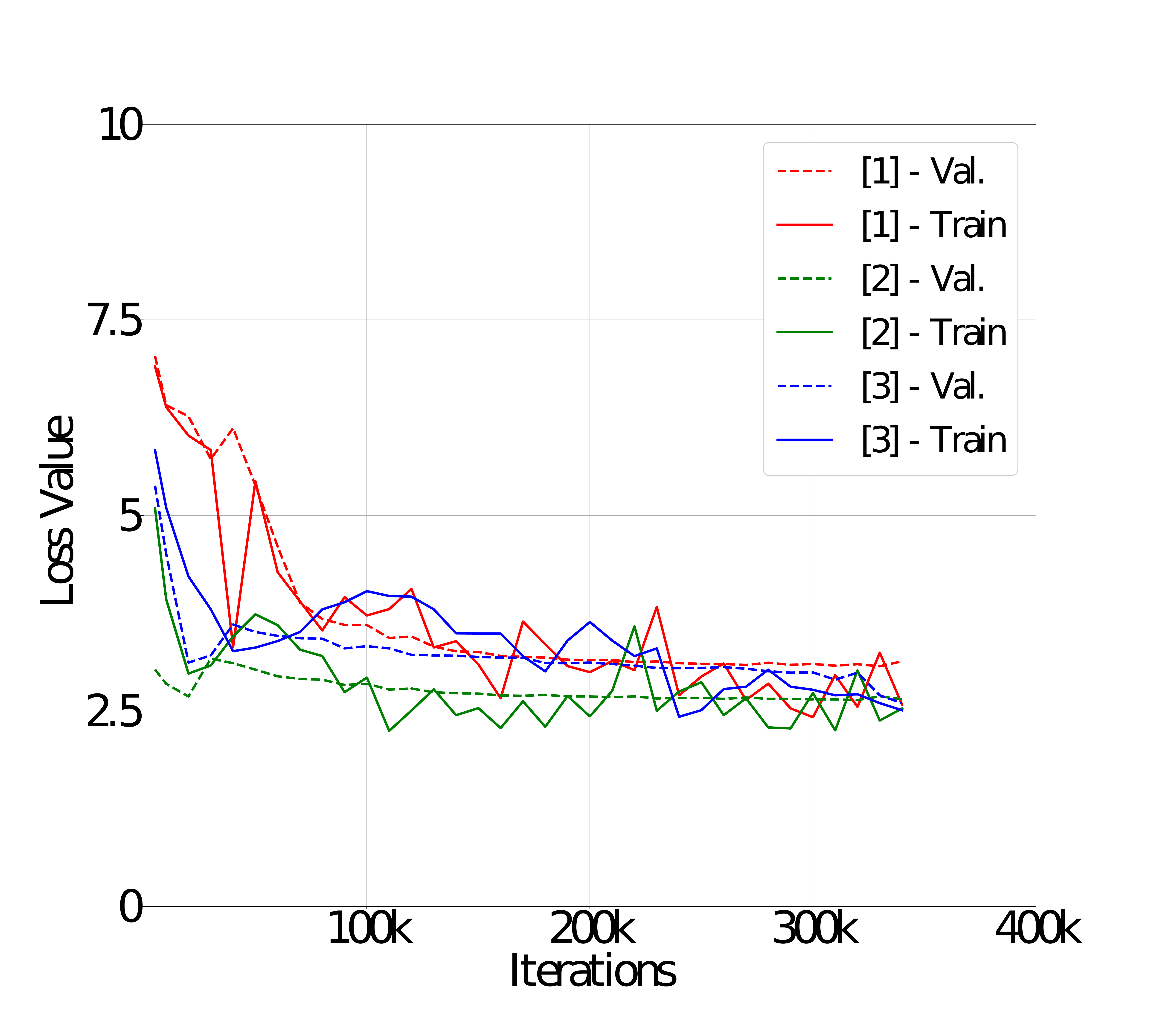}
	\caption{Validation and training loss values v.s. the number of iterations for summarization models}
	\label{fig_lossvals}
\end{figure}
\\
Figure \ref{fig_lossvals} shows the loss on the training set and validation set for as a function of number of iterations for the summarization models for 340,000 iterations (19 epochs). The results of summarization are compared for one case v.s. its ground truth for the three summarization models in Table \ref{tab_modelOutputs}. As it can be seen the summary generated by model [1] contains $<unk>$ instead of the word \textit{mysak} in the original summary. However, due to having attention and the pointer-generator mechanism model [2] has replaced the $<unk>$ with the proper word from the source text. However, summary of model [2] has repeated a sentence twice. The summary generated by the pointer-generator together with the coverage mechanism not only could have overcome the $<unk>$ problem but also does not have repetition in the generated summary and gives a nice summary pretty close to the reference summary. The summary generated by the transformer model can only capture some keywords but does not convey the grasp of summary very well.\\

\begin{table}[]
\resizebox{\columnwidth}{!}{%
\begin{tabular}{|c|l|}
\hline
{\color[HTML]{3166FF} \textbf{Reference}} & \begin{tabular}[c]{@{}l@{}}once a super typhoon , maysak is now a tropical storm with 70 mph winds .\\ it could still cause flooding , landslides and other problems in the philippines .\end{tabular}                                                                                                   \\ \hline
\textbf{Model [1]}                         & \begin{tabular}[c]{@{}l@{}}{[}UNK{]} gained super typhoon status thanks to its sustained 150 mph winds. \\ it 's now classified as a tropical storm. \\ it 's expected to make landfall sunday on the southeastern coast of {[}UNK{]} province .\end{tabular}                                            \\ \hline
\textbf{Model [2]}     & \begin{tabular}[c]{@{}l@{}}tropical storm maysak approached the asian island nation saturday .\\ it's now classified as a tropical storm , according to the philippine national weather service .\\ it's now classified as a tropical storm , according to the philippine weather service .\end{tabular} \\ \hline
\textbf{Model [3]}          & \begin{tabular}[c]{@{}l@{}}just a few days ago , maysak gained super typhoon status thanks to its sustained 150 mph winds .\\ it 's now classified as a tropical storm , according to the philippine national weather service .\end{tabular}                                                             \\ \hline
\textbf{Model [4]}          & \begin{tabular}[c]{@{}l@{}}super typhoon could weaken . new jersey , but it will . \\philippine ocean strength . at least 132 people are injured ,  including 18  .\end{tabular}                                                             \\ \hline
\end{tabular}
}
\caption{Comparison of the generated summary using the summarization models v.s. the ground truth}
\label{tab_modelOutputs}
\end{table}

\textbf{Fake News Detection Subsequent to Summarization} 

In this part, we use the best summarization model that we have trained on the summarization dataset in order to create summaries of a fake news detection dataset. We will build a fake news detection model and we investigate its performance when the input is the original news text, the news headline, and the summarized news text generated by our summarization model. Basically, We use our text summarizing model as a feature generator for a fake news classification model. In fake news classification the article content contains much more information than the article headline and due to this a fake news classifier performs better on article contents than on article headlines.
\begin{figure}[!ht]
\centering 
\includegraphics[width=1\textwidth]{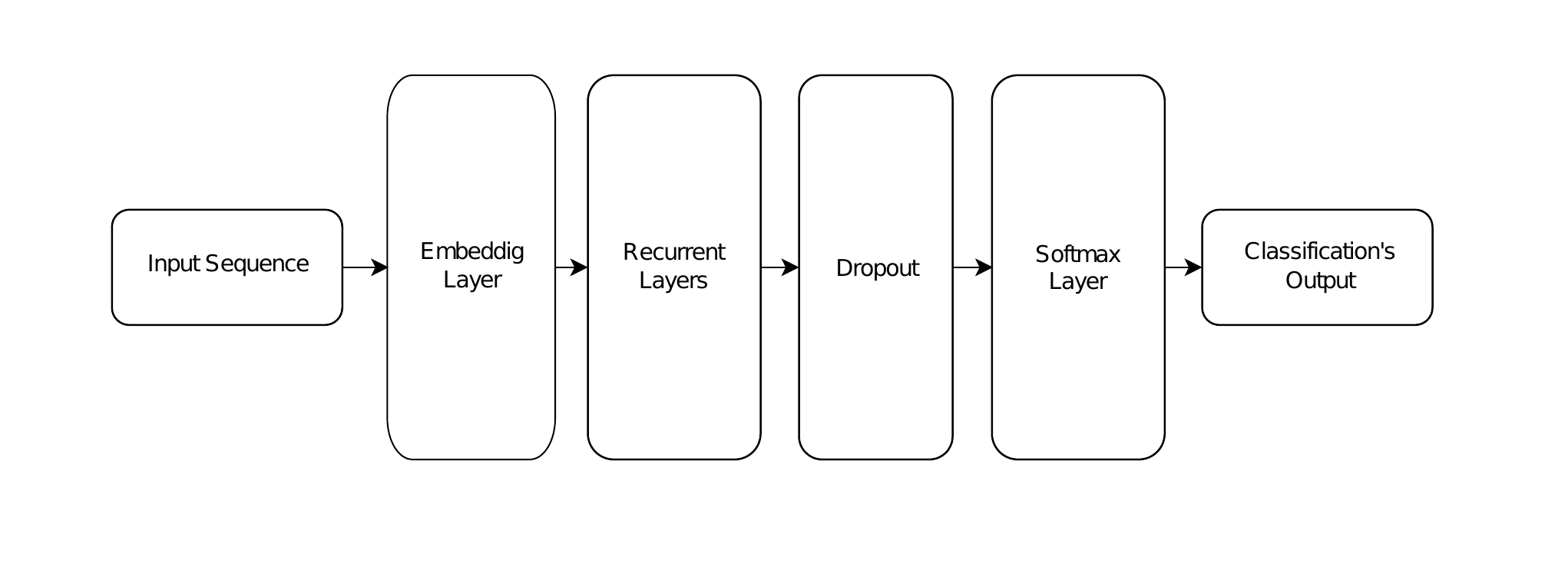}
	\caption{Fake news classification architecture}
	\label{fig_LSTM_Fake}
\end{figure}
\begin{figure}
     \centering
     \begin{subfigure}[b]{0.3\textwidth}
         \centering
         \includegraphics[width=\textwidth]{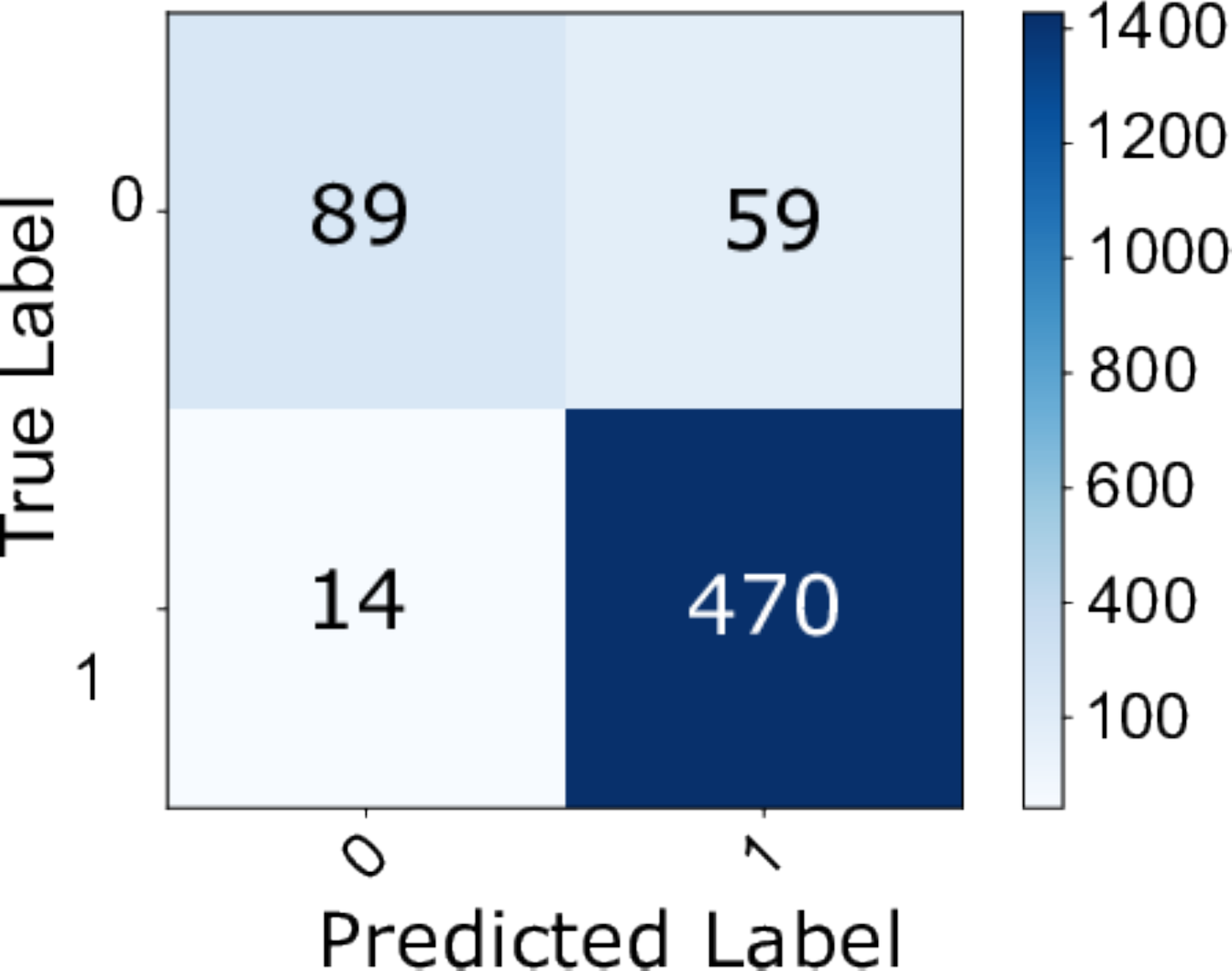}
         \caption{Full body text}
         \label{fig:y equals zz}
     \end{subfigure}
     \hfill
     \begin{subfigure}[b]{0.3\textwidth}
         \centering
         \includegraphics[width=\textwidth]{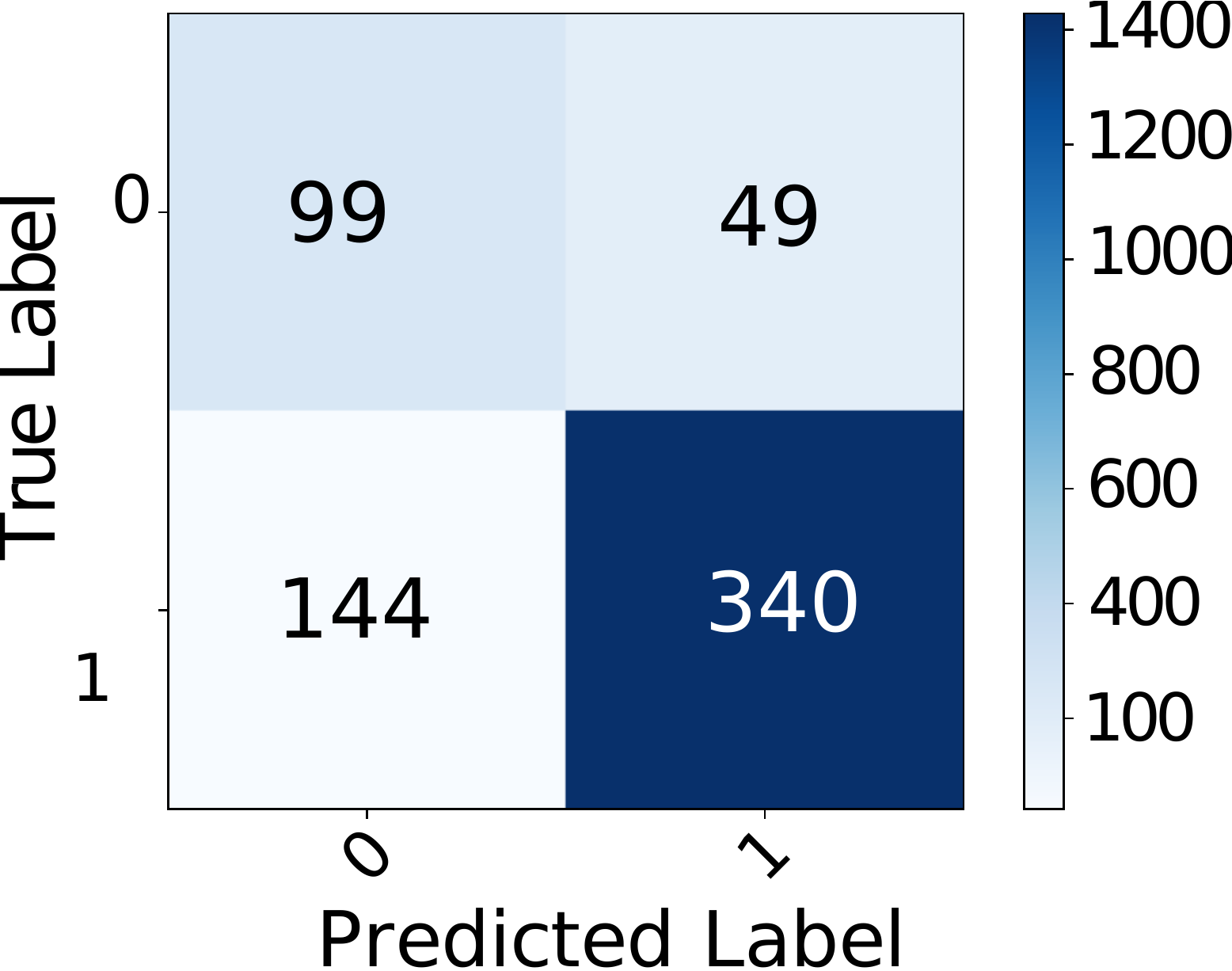}
         \caption{Headline text}
         \label{fig:three sin x}
     \end{subfigure}
     \hfill
      \begin{subfigure}[b]{0.3\textwidth}
         \centering
         \includegraphics[width=\textwidth]{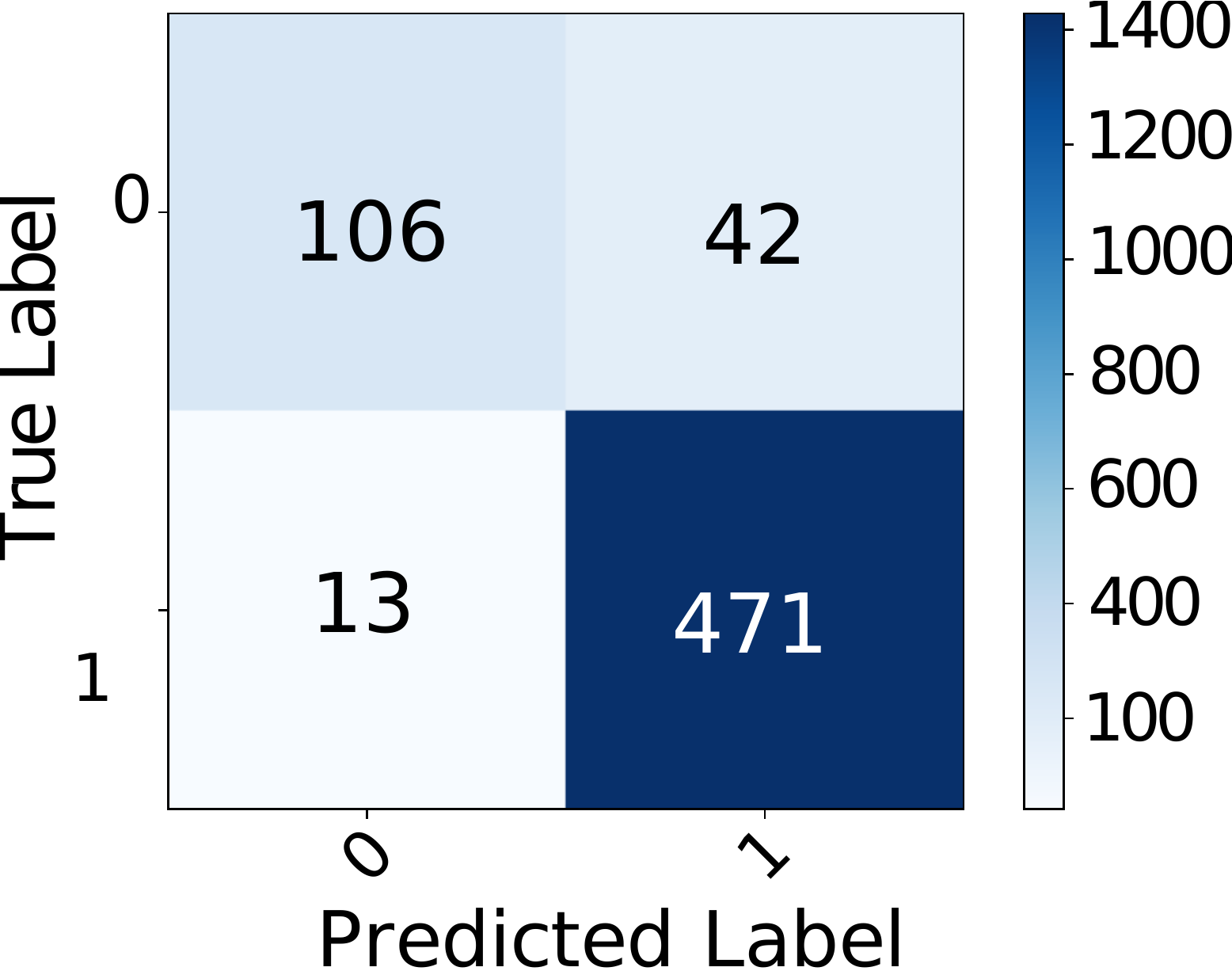}
         \caption{Summary text}
         \label{fig:y equals x}
     \end{subfigure}
        \caption{Confusion matrix for test set of fake news detection task  using three different input features}
        \label{fig_confMatrices}
\end{figure}
\renewcommand{\arraystretch}{1.05}
\begin{table}[h!]
\centering
\resizebox{0.95\columnwidth}{!}{
\begin{tabular}{c|c|ccc|cccc|}
\cline{1-9}
\multicolumn{1}{|c|}{\textbf{Input features}} & \multicolumn{1}{c|}{\textbf{Exp. \#}} & \multicolumn{1}{c|}{\textbf{Cells}} & \multicolumn{1}{c|}{\textbf{Size}} & \multicolumn{1}{c|}{\textbf{Drop-out}} & \multicolumn{1}{c|}{\textbf{train loss}} & \multicolumn{1}{c|}{\textbf{train acc. \%}} & \multicolumn{1}{c|}{\textbf{valid loss}} & \multicolumn{1}{c|}{\textbf{valid acc. \%}} \\ \hline
\multicolumn{1}{|c|}{\multirow{8}{*}{Full body text}} & 1 & \textbf{LSTM} & \textbf{64} & \textbf{0.2} &  0.081&97.4&0.12&\textbf{92.1} \\
\multicolumn{1}{|c|}{} & 2 & LSTM & 64 & 0.5 & 0.143&93.7&0.167&91.5  \\
\multicolumn{1}{|c|}{} & 3 & LSTM & 128 & 0.2 & 0.076&97.2&0.178&91.4  \\
\multicolumn{1}{|c|}{} & 4 & LSTM & 128 & 0.5 &  0.141&94.5&0.224&90.2  \\
\multicolumn{1}{|c|}{} & 5 & Bi-LSTM & 64 & 0.2 &  0.025&99.1&0.129&90.9  \\
\multicolumn{1}{|c|}{} & 6 & Bi-LSTM & 64 & 0.5 &  0.08&97.1&0.128&91.2 \\
\multicolumn{1}{|c|}{} & 7 & Bi-LSTM & 128 & 0.2 & 0.026&99.2&0.111&89.7  \\
\multicolumn{1}{|c|}{} & 8 & Bi-LSTM & 128 & 0.5 &  0.0741&97.3&0.113&88.6  \\ \cline{1-1}
\hline
\hline
\multicolumn{1}{|c|}{\multirow{8}{*}{Headline text}} & 1 & LSTM & 64 & 0.2 & 0.121 & 95.3 & 0.241 & 91.6 \\
\multicolumn{1}{|c|}{} & 2 & LSTM & 64 & 0.5 & 0.157 & 93.2 & 0.215 & 91.3 \\
\multicolumn{1}{|c|}{} & 3 & LSTM & 128 & 0.2 & 0.099 & 95.9 & 0.227 & 91.7 \\
\multicolumn{1}{|c|}{} & 4 & LSTM & 128 & 0.5 & 0.156 & 93.6 & 0.221 & 91.0 \\
\multicolumn{1}{|c|}{} & 5 & \textbf{Bi-LSTM} & \textbf{64} & \textbf{0.2} & 0.103 & 95.6 & 0.229 & \textbf{91.8} \\
\multicolumn{1}{|c|}{} & 6 & Bi-LSTM & 64 & 0.5 & 0.154 & 93.5 & 0.219 & 91.0 \\
\multicolumn{1}{|c|}{} & 7 & Bi-LSTM & 128 & 0.2 & 0.106 & 95.7 & 0.239 & 91.5 \\
\multicolumn{1}{|c|}{} & 8 & Bi-LSTM & 128 & 0.5 & 0.158 & 93.5 & 0.217 & 91.1 \\ \cline{1-1}
\hline
\hline
\multicolumn{1}{|c|}{\multirow{8}{*}{Summary text}} & 1 & LSTM & 64 & 0.2 & 0.074 & 97.3 & 0.291 & 92.1 \\
\multicolumn{1}{|c|}{} & 2 & LSTM & 64 & 0.5 & 0.146 & 94.5 & 0.231 & 92.2 \\
\multicolumn{1}{|c|}{} & 3 & LSTM & 128 & 0.2 & 0.083 & 97 & 0.247 & 92.3 \\
\multicolumn{1}{|c|}{} & 4 & LSTM & 128 & 0.5 & 0.139 & 94.6 & 0.201 & 91.3 \\
\multicolumn{1}{|c|}{} & 5 & Bi-LSTM & 64 & 0.2 & 0.078 & 97.1 & 0.291 & 91.9 \\
\multicolumn{1}{|c|}{} & 6 & Bi-LSTM & 64 & 0.5 & 0.152 & 94.1 & 0.246 & 91.6 \\
\multicolumn{1}{|c|}{} & 7 & \textbf{Bi-LSTM} & \textbf{128} & \textbf{0.2} & 0.079 & 97.1 & 0.221 & \underline{\textbf{93.1}} \\
\multicolumn{1}{|c|}{} & 8 & Bi-LSTM & 128 & 0.5 & 0.146 & 94.5 & 0.242 & 91.8 \\ \cline{1-1}
\hline
\end{tabular}
}
\caption{Experiments on the fake news detection}
\label{table_fakeNewsExperiments}
\end{table}

\renewcommand{\arraystretch}{1.1}
\begin{table}[h!]
\centering
\resizebox{0.6\columnwidth}{!}{
\begin{tabular}{|c|c|c|}
\hline
\textbf{Input Features}  & \textbf{Accuracy \%} & \textbf{Average Length (in words)} \\ \hline
Full body text  & 92    & 10.51       \\ \hline
Headline text   & 91    & 387.89      \\ \hline
Summary text    & 93    & 20.41       \\ \hline
\end{tabular}
}
\caption{Fake news classifier results}
\label{tab_fakenews_summarizedResults}
\end{table}
For fake news classification, we use a fake news dataset with headlines and article content provided by George McIntire \footnote{https://www.datasciencecentral.com/profiles/blogs/on-building-a-fake-news-classification-model}. The dataset contains 3164 fake news articles and 3171 real articles (i.e. a balanced dataset) on politics from a wide range of news sources. We shuffle the data and use 80\% of it for training, 10\% of it for validation, and 10\% for testing, and also do 5-fold cross validation. \\
We build a Long-short-term-memory (LSTM) network together with an Embedding Layer as shown in Figure \ref{fig_LSTM_Fake}. Table \ref{table_fakeNewsExperiments} shows our hyperparameter studies for fake news classification and Table \ref{tab_fakenews_summarizedResults} shows the final test accuracies, using the three input features of full body text, headline text, and generated summary texts by our summarization models. As it can be seen in this table the best model using the body text as input features perform better than headline text as input. Furthermore, it's worth noting that the summary text as input feature leads to an even higher accuracy compared to the full body text as input feature. This finding shows that summarization model serves as a feature generator for fake news detection task which actually increases its accuracy. Also, this summarization model can also serve as a headline generator for the news articles as an automatic approach.
\section{Conclusion}
As we showed in section \ref{sec_results} the pointer-generator architecture with attention and coverage mechanisms led to the highest accuracies and could overcome the problems common in abstractive text summarization such as out-of-vocabulary words and repetition. Furthermore, as shown in section \ref{sec_results} a text summarizing model can successfully be applied as a feature generator prior to classification tasks such as fake news classification and increase the accuracy of those tasks.

\bibliographystyle{plainnat} 


\clearpage



\section*{References}

\begin{thebibliography}{9}

\bibitem[Esmaeilzadeh(2018 a)]{esmaeilzadeh_alz3} 
\newblock Soheil Esmaeilzadeh, Ouassim Khebzegga, and Mehrad Moradshahi.
\newblock \textit{Clinical Parameters Prediction for Gait Disorder Recognition}. 
\newblock arXiv:1806.04627. 
\newblock 2018. 
\newblock \url{https://arxiv.org/abs/1806.04627}.

\bibitem[Esmaeilzadeh(2018 b)]{esmaeilzadeh_alz} 
\newblock Soheil Esmaeilzadeh, Dimitrios Ioannis Belivanis, Kilian M. Pohl, and Ehsan Adeli.
\newblock \textit{End-to-end Alzheimer’s disease diagnosis and biomarker identification}. 
\newblock Machine Learning in Medical Imaging. 
\newblock MLMI 2018. 
\newblock pp  337-345. 
\newblock vol 11046. 
\newblock \url{https://doi.org/10.1007/978-3-030-00919-9_39}.

\bibitem[Esmaeilzadeh(2018 c)]{esmaeilzadeh_alz2} 
\newblock Soheil Esmaeilzadeh, Yao Yang, and Ehsan Adeli.
\newblock \textit{End-to-End Parkinson Disease Diagnosis using Brain MR-Images by 3D-CNN}.  arXiv:1806.05233. 2018. \newblock \url{https://arxiv.org/abs/1806.05233}.

\bibitem[Cheng(2019)]{Cheng2019} 
Pengxiang Cheng, and Katrin Erk
\textit{Attending to Entities for Better Text Understanding}. 	arXiv:1911.04361. 2019. \url{https://arxiv.org/abs/1911.04361}

\bibitem[Liu(2018)]{Liu2018} 
Hui Liu, Qingyu Yin, and William Yang Wang
\textit{Towards Explainable NLP: A Generative Explanation Framework for Text Classification}. arXiv:1811.00196. 2018 \url{https://arxiv.org/abs/1811.00196}

\bibitem[Allahyari(2017)]{Allahyari} 
Mehdi Allahyari, Seyedamin Pouriyeh, Mehdi Assefi, Saeid Safaei, Elizabeth D. Trippe, Juan B. Gutierrez, and Krys Kochut.
\textit{Text Summarization Techniques : A Brief Survey}. arXiv:1707.02268. 2017. \url{https://arxiv.org/abs/1707.02268}

\bibitem[Dorr(2003)]{Dorr2007} 
Bonnie Dorr, David Zajic, and Richard Schwartz.
\textit{Hedge Trimmer: A Parse-and-Trim Approach to Headline Generation}. 
Proceedings of the HLT-NAACL 03 Text Summarization Workshop. pp. 1-8, 2003. \url{http://doi.org/10.3115/1119467.1119468}

\bibitem[Nallapati(2016)]{Nallapati2016} 
Ramesh Nallapati, Feifei Zhai, and Bowen Zhou.
\textit{SummaRuNNer: A Recurrent Neural Network based Sequence Model for Extractive Summarization of Documents}.
arXiv:1611.04230. 2016. \url{https://arxiv.org/abs/1611.04230}

\bibitem[Khatri(2018)]{Khatri2018} 
Chandra Khatri, Gyanit Singh, and Nish Parikh.
\textit{Abstractive and Extractive Text Summarization using Document Context Vector and Recurrent Neural Networks}.
arXiv:1807.08000. 2018. \url{https://arxiv.org/abs/1807.08000}

\bibitem[Gao(2018)]{Gao} 
Shen Gao, Xiuying Chen, Piji Li, Zhaochun Ren, Lidong Bing, Dongyan Zhao, and Rui Yan.
\textit{Abstractive Text Summarization by Incorporating Reader Comments}.
arXiv:1812.05407. 2018. \url{https://arxiv.org/abs/1812.05407}

\bibitem[Bahdanau(2014)]{Bahdanau} 
Dzmitry Bahdanau, Kyunghyun Cho, and Yoshua Bengio.
\textit{Neural Machine Translation by Jointly Learning to Align and Translate}.
arXiv:1409.0473. 2014. \url{https://arxiv.org/abs/1409.0473}

\bibitem[Nallapati(2016)]{Nallapati2016a} 
Ramesh Nallapati, Bing Xiang and Bowen Zhou.
\textit{Sequence-to-Sequence RNNs for Text Summarization}. ICLR 2016. 

\bibitem[Hermann(2015)]{Moritz} 
Karl Moritz Hermann, Tomas Kocisky, Edward Grefenstette, Lasse Espeholt, Will Kay, Mustafa Suleyman, and Phil Blunsom.
\textit{Teaching Machines to Read and Comprehend}. Advances in Neural Information Processing Systems. NIPS 2015. 

\bibitem[See(2017)]{See2013} 
Abigail See, Peter J. Liu, and Christopher D. Manning.
\textit{Get To The Point: Summarization with Pointer-Generator Networks}. arXiv:1704.04368. 2017. \url{https://arxiv.org/abs/1704.04368}.

\bibitem[Vaswani(2017)]{Vaswani2017} 
Ashish Vaswani, Noam Shazeer, Niki Parmar, Jakob Uszkoreit, Llion Jones, Aidan N. Gomez, Lukasz Kaiser, and Illia Polosukhin.
\textit{Attention Is All You Need}. arXiv:1706.03762. 2017. \url{https://arxiv.org/abs/1706.03762}.

\bibitem[Tu(2016)]{Tu2016} 
Zhaopeng Tu, Zhengdong Lu, Yang Liu, Xiaohua Liu, and Hang Li.
\textit{Modeling Coverage for Neural Machine Translation}.
arXiv:1601.04811. 2016. \url{https://arxiv.org/abs/1601.04811
}

\bibitem[Chen(2016)]{Chen} 
Danqi Chen, Jason Bolton, and Christopher D. Manning.
\textit{A Thorough Examination of the CNN/Daily Mail Reading Comprehension Task}. Proceedings of the 54th Annual Meeting of the Association for Computational Linguistics. ACL 2016. \url{http://doi.org/10.18653/v1/P16-1223}

\bibitem[Koupaee(2018)]{Wang} 
Mahnaz Koupaee, and William Yang Wang.
\textit{WikiHow: A Large Scale Text Summarization Dataset}. 	arXiv:1810.09305. 2018. \url{https://arxiv.org/abs/1810.09305}

\bibitem[Lin(2004)]{Lin2001} 
Chin Yew Lin, and Marina Rey.
\textit{ROUGE : A Package for Automatic Evaluation of Summaries}. Text Summarization Branches Out. ACL 2004.




\end{thebibliography}

\end{document}